\documentclass[runningheads]{llncs}
\usepackage{amsmath}
\usepackage[T1]{fontenc}
\usepackage{algorithm}
\usepackage{algpseudocode}
\usepackage{array, booktabs}
\usepackage[english]{babel}
\usepackage[square,numbers]{natbib}
\usepackage{hyperref}
\hypersetup{
  colorlinks   = true, 
  urlcolor     = blue, 
  linkcolor    = blue, 
  citecolor   = green 
}
\usepackage{graphicx}
\begin{document}
\title{Panoptic Segmentation of Environmental UAV Images : Litter Beach }
%
%
\author{Ousmane YOUME\inst{1} \and
Jean Marie Dembele\inst{1,2}\and
Eugene C. Ezin\inst{3} \and Christophe Cambier\inst{2}}
\authorrunning{O. Youme et al.}
%
\institute{Gaston Berger University, Saint-Louis, Senegal 
\email{youme.ousmane1@ugb.edu.sn} \and
UMI 209 UMMISCO, Sorbonne University IRD
\and
IMSP \& IFRI, University of Abomey-Calavi, Abomey-Calavi, Benin}
\maketitle              
\begin{abstract}
Convolutional neural networks (CNN) have been used efficiently in several fields, including environmental challenges. In fact, CNN can help with the monitoring of marine litter, which has become a worldwide problem. UAVs have higher resolution and are more adaptable in local areas than satellite images, making it easier to find and count trash. Since the sand is heterogeneous, a basic CNN model encounters plenty of inferences caused by reflections of sand color, human footsteps, shadows, algae present, dunes, holes, and tire tracks. For these types of images, other CNN models, such as CNN-based segmentation methods, may be more appropriate. In this paper, we use an instance-based segmentation method and a panoptic segmentation method that show good accuracy with just a few samples. The model is more robust and less sensitive to disturbances, especially in panoptic segmentation.
\end{abstract}

\section{Introduction}\label{sec1}

Marine pollution is a phenomenon that is growing every year at an unprecedented rate. Each year, more than 10 million tons of waste are thrown onto beaches and oceans, $80\%$ of which comes from land and $20\%$ from maritime activities.
They are made up of $80\%$ plastic and their destruction constitutes a problem because the degradation of the polymers leads to microparticles after fragmentation. \cite{ref_article1}  Waste on the coast, in addition to being an important ecological factor, harms tourism, the economy, and the health of ecosystem species. Therefore, it is essential to find techniques to automatically detect and list the trash on the coast to make treatment easier.\cite{ref_article2}

Traditional counting methods have been suggested, such as the OSPAR 2010 guidelines \cite{ref_article3}, which call for a $100meter$  sweep of the beach to collect trash larger than $2.5cm$. Then, it would take $2$–$5$ collectors $3$–$5$ hours to sort the trash and measure the quantity. With the progress of technology, it has recently been shown \cite{ref_article4} that aerial images and machine learning methods can be used to monitor and find waste. Some work has been done in this sense. Cecilia et al. \cite{ref_article4} use machine learning to automatically find trash in the ocean and describe it. Three (3) random forests are built: two for binary true or false predictions and the third for predicting the species of true boxes: plastic, capsule, etc. Gonzalez et al. \cite{ref_article5} also use random forest classification to find marine litter. The input images are changed upstream by adding three types of color chains to the RGB: the HSV, the CIE-Lab, and the YCbCr. This creates an input vector with $12$ features. Marine litter is detected with an accuracy of F-Score of $76\%$ on the beach and an accuracy of F-Score of $55\%$ on the dunes. 
At the ImageNet Large-Scale Visual Recognition Challenge (ILSVRC), deep learning methods were the best. However, basic CNN models do not work well for task detection in high-resolution images of the environment taken by satellite or UAV \cite{ref_article6}. These models show a lot of inferences in a dense area. The task is hard because the color of the sand reflects off objects, footsteps, shadows, algae, and other things that make it hard to find litter. In this paper, we show that segmentation methods work better for this kind of image of the environment. Two models were chosen for experimentation and compared in this work. One method is based on instance segmentation, and another is based on panoptic segmentation. Both are used on a UAV-collected dataset of trash on the coast.

\section{Related works }\label{sec2}
In the S.O.T.A, a few works are noted in the application of CNN for the detection of coastal wastes. Fallati et al. \cite{ref_article7}  employ deep learning, specifically CNN. They use commercial plastic detection software to generate numerical scores on two different sites ($44\%$ and $78\%$ respectively).
 Papakonstantinou et al \cite{ref_article8} proposed a UAS data acquisition and annotation protocol combined with deep learning techniques for automatic detection and mapping of ML concentrations in the coastal zone. Testing the generalization ability of convolutional neural networks on an unseen dataset, they found that the VGG19 \cite{ref_article9}  network for more than $15.000$ samples yielded an overall accuracy of $77.6\%$ and a F-Score of $77.42\%$. Wolf et al. (2021) \cite{ref_article10} demonstrate that a basic convolutional neural network (CNN) can identify and count floating and shoreline-dumped plastic litter. The Aquatic Plastic Waste Detection, Classification, and Quantification System (APLASTIC-Q) was developed and trained using high-resolution geospatial images taken during aerial surveys in Cambodia, a local environment that is not complex and varied. The machine learning parts of APLASTIC-Q are the plastic waste detector (PLD-CNN) and the plastic waste quantifier (PLQ-CNN). The PLD-CNN successfully classified the targets into water, sand, vegetation, and plastic waste with an accuracy of $83\%$. YH Liao 2022 \cite{ref_article11}  proposes a marine litter detection system that uses drones. The goal is to use drones instead of people to find trash in the ocean and give government agencies information about pollution in real time. This study used the Internet and computer-to-computer communication. Images of marine litter were provided to train a modified YOLO \cite{ref_article12}  model (You Look Only Once). With an accuracy of around of $70\%$ at most of the models cited above, it should be recalled that these models are trained on images taken from close-up and non-heterogeneous areas. Therefore, we cannot guarantee inference in dense or diverse areas. In this sense, Fallati et al. affirm \cite{ref_article7} that a basic CNN architecture has a high number of false positives caused by sand dunes, shadows, footsteps, algae, and other things leaving traces that mess with the model . The purpose of this paper is to find a more robust and sensitive model adapted for this type of area.

\section{Background}\label{sec3}
Contemporary methods such as large-scale VGG19 \cite{ref_article9}, and Xception \cite{ref_article13} have been proposed for efficient object detection in videos or images taken on land. But it's not always easy to find a suitable model for a problem when there aren't any ready-made solutions. Also, these methods don't seem to work well with high-resolution images from a drone or a satellite. For this kind of data, segmentation methods are more commonly used in the literature. In the next section, we present the segmentation methods. 

\subsection{Segmentation Methods}\label{subsec31}

\subsubsection{Semantics Segmentation}\label{subsec32}
Semantic segmentation is a pixel-level classification of the image. An embedding of the boundary changes between objects of different colors is made. Its major development starts with the proposal of the FCN \cite{ref_article16}. After a sequence of convolution and reduction, the FCN turns the input image into an output of the same size. The output is then given an up-convolution, which sorts each pixel according to a label. The Unet \cite{ref_article15} based on FCN, is a segmentation method composed of two parallel paths. The first path reduces dimension, including $conv(3\times3)$ and $max pooling(2\times2)$. With pixel classification, the image dimensions are put back together by following an extended path called up-convolution. HRNET \cite{ref_article17}, inspired by Unet \cite{ref_article15}, is made up of four parallel branches, each with a different reduction dimension. ASP \cite{ref_article18} uses HRNET \cite{ref_article17} and uses the first two branches of adaptive spatial pooling to increase the gain in the extraction feature at different levels. There are other models based on spatial pyramid pooling, such as PSPNET\cite{ref_article19}, DeepLab\cite{ref_article14}, etc.

\subsubsection{Instance Segmentation}\label{subsec33}
 Different from semantic segmentation, instance-based segmentation, in addition to classification at the pixel level, adds object detection techniques. This combination allows for more robustness on object detection tasks with a distinction between instances of the same label. A unique instance ID is assigned to each object instance. This enables simple object quantity counting.  We distinguish two (2) main methodologies: the bottom-up technique, in which segmentation is used first and then object detection is applied to the proposed output (Deeperlab \cite{ref_article20}), and the top-down method, in which both techniques are used in parallel and the results of these two prediction branches are merged. Top-Down is divided into two categories: two-stage methods that integrate the region proposal with parallel mask branches i.e, Mask R-CNN \cite{ref_article21}, EfficientNet \cite{ref_article22}, etc., and one-stage methods that eliminate the region proposal, i.e, Polarmask \cite{ref_article23}. 

\subsubsection{Panoptics Segmentation}\label{subsec34}

Panoptic segmentation combines semantics and instance segmentation techniques. The panoptic identifies the expected output objects and fuses them with mask branch pixel label classification. The rest of the image, including the background, is then separated at the pixel level using semantic segmentation techniques. Most panoptic segmentation models add parallel paths for semantic segmentation to existing per-instance segmentation models. As a result, they include bottom-up instance segmentation models such as Panoptic DeepLab \cite{ref_article24},  two-stage methods such as EfficientPS \cite{ref_article25}, one-stage methods such as FPSNet \cite{ref_article26}, and also some single-path methods.

\section{Methodologies}\label{sec4}
\subsection{Area of Study}\label{subsecsec41}
The study area is located on the coast of Dakar, Senegal, in West Africa. Senegal is a country that registers a lot of tourists and must be favorable for the protection of the environment. On the other hand, some of its places have a lot of waste, especially on the beaches along the coastline. Tavares et al \cite{ref_article27} in a study of two locations, Dakar and Mbour, found an average density of debris at least $10 cm$ deep of $48.75$ debris per m2 and 1.95 debris per $m^2$ on the surface, $95\%$ of which is plastic.  As a result, the establishment of a monitoring system is critical in order to have information on the location and density of waste and thus contain in order to effectively eliminate marine pollution. 
The place selected is located north of Dakar on the coastline from Dakar to the region of Saint-Louis at an altitude of 14 ° 48' north and 17 ° 18 longitude. The UAV flew at altitudes of 5, 10, and 30 meters over a distance of 300 meters at a 90-degree angle. Given the images taken, the altitude of 10 meters is more satisfactory in terms of coverage, image resolution, and visibility of debris. For scale variation, we will therefore use the images captured at $10$ and $30 m$ to train our model. 

\subsection{Data Acquisition}\label{subsecsec42}

For the acquisition of necessary data, we use the DJI Mavic PRO drone, which has a high resolution with its L1D-20c RGB color camera, taking images of $5472\times3648$ pixels. The high resolution is essential for the performance of our model. In addition, a UAV is easily manipulated and contains metadata such as shutter speed, aperture, ISO, and GPS coordinates (latitude, longitude, and altitude) that can be read with the \textit{exiftool} software. Given the intensity of the light, which exacerbates the problem with shadows and the whiteness of the sand, each image is subjected to an RGB color attenuation algorithm. The curve of the histogram with the parameters of contrast and brightness on each image is taken, and then they are automatically optimized by defining the optimal alpha (contrast) and beta (brightness), knowing that: 
\begin{align}
f(i,j)=\alpha \times g(i,j) + \beta, \label{eq1}
\end{align}
i and j are the pixel coordinates, f is the new image, and g is the old. When the frequency of colors is less than a certain threshold value (e.g. 1), the cumulative distribution is computed, and the right and left ends of the histogram are cut off. This gives the minimum and maximum ranges. Here is a visualization of the histogram before (orange) and after the clipping (blue). Notice that the most "interesting" sections of the image are more pronounced after cropping.

To calculate the alpha, we take the minimum and maximum gray ranges after clipping and divide them by the desired output range of 255. 
\begin{align}
\alpha=  255 / (maximumgray - minimumgray).   \label{eq2}
\end{align}
To calculate the beta, we plug it into the formula where
\begin{align}
g(i, j)=0,     \label{eq3}
\\
f(i, j) = minimumgray, \label{eq34}
\\
g(i, j) = \alpha \times f(i, j) + \beta.  \label{eq5}
\end{align}
which, after being solved, results in:
 \begin{align}
 \beta= -minimumgray \times \alpha
 \end{align}
 
 \begin{figure}[H]
    \centering
    \includegraphics[width=7.5cm, height=5.5cm]{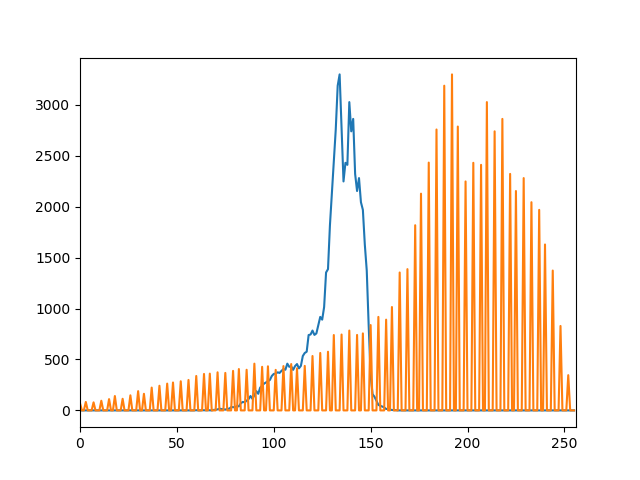}.
    \caption{Difference of variation in images before (orange) and after processing (blue).}
    \label{fig2}
\end{figure}
This equation gives alpha and beta to apply to each image. Trash in images obtained after transformation is more visible and differentiates the background, which is the sand. The transformed images are divided into a grid of images of size $600 \times 600$ adapted to the inputs of the chosen algorithm. This corresponds for each image transformed into $72$ small images. A final database is obtained for annotation and training the model. 

\section{Algorithms Selected}\label{sec5}
Segmentation architectures can have many different parts, such as anchor bases, bounding boxes, and non-maximum suppression. There are four types of processes that use these mechanisms: top-down (one stage and two stages), bottom-up, single-path, and merged methods. 
The segmentation by instance model: Mask R-CNN, and the panoptic model: Panoptic DeepLab are chosen for experimentation based on their specificity. They rank well based on their performance on the Coco Val dataset, which is closest to our local contexts. We recall that experimentation is more related to the suitable type of segmentation, regardless of the model, than to achieving high precision. Mask R-CNN \cite{ref_article22} is a classical object detection algorithm with instance-based segmentation. Mask R-CNN \cite{ref_article22} extends Fast R-CNN \cite{ref_article28} with two parallel branches: the first one for classification or regression of bounding boxes and the second one for prediction of masks applied to each ROI. 
Panoptic-DeepLab \cite{ref_article25} employs a central keypoint for class-agnostic instance detection, in contrast to DeeperLab \cite{ref_article20}. As a result, Panoptic-DeepLab predicts three outputs: (1) semantic segmentation, (2) instance center heatmap, and (3) instance center regression. First, class-agnostic instance segmentation is made by putting the predicted foreground pixels together with the predicted instance centers that are closest to them. When the class-agnostic instance segmentation and the semantic segmentation are put together, the final panoptic segmentation is made.
 \begin{figure}[H]
    \centering
    \includegraphics[width=7.5cm, height=4.5cm]{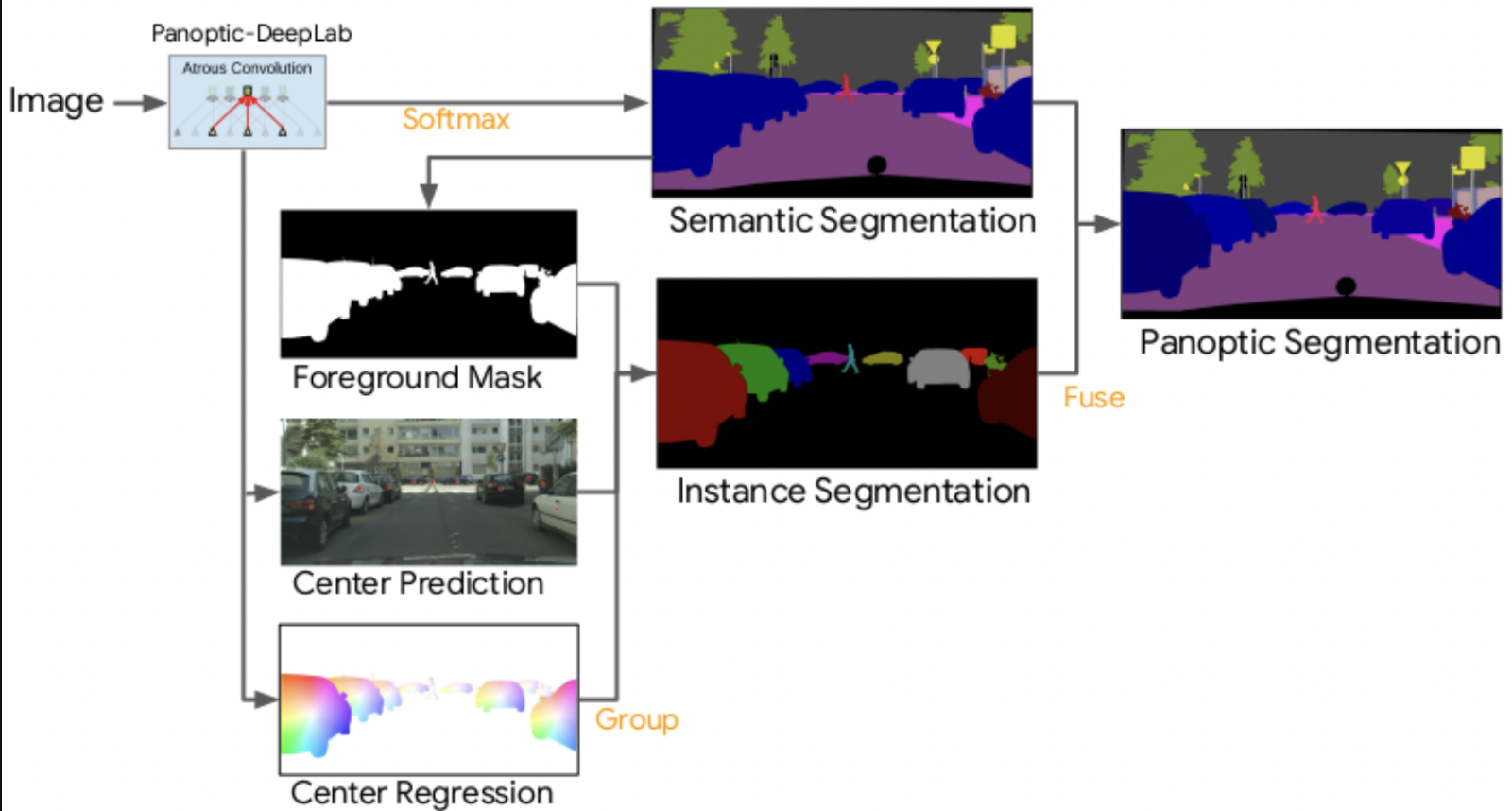}.
    \caption{Panoptic Deeplab architecture images from the original paper.\cite{ref_article24}}
    \label{fig2}
\end{figure}
The loss function is calculate as follow:
\begin{align}
L = \lambda_1( L_c + L_b + L_m)+ \lambda_s \times L_s 
\end{align}
$L_c$ is a softmax cross-entropy classification loss function, $L_b$ is a regression loss smooth, and $L_1$ is the mask loss function. $lambda_1$ and $lambda_s$ are set to zero or one depending on the training stage, either instance or semantic.  
\section{Experimentation }\label{sec6}
\subsection{Litter Dataset}\label{subsec61}
The litter dataset is made up of $1500$ images of $600\times600$ pixels after data collection by UAV and image pre-processing. For the labels "Litter" and "Algae," a total of $10$ worth of objects is calculated per class. Images are labeled manually with VGG Image Annotator. Finally, a JSON file containing polygons representing the positions of objects in the image is exported. Litter Dataset is split in three (3) datasets: training set for $55\%$, test set $35\%$, and validation set $10\%$. As metrics, Average Precision, PS (precision in small objects), PL (precision in large objects), and AR (average recall) are retained. 

\subsection{Parameters Initialization}\label{subsec62}
For parameter initialization, all models share the same optimization parameters. The learning rate is set to $0.001$, and the batch size is set to $2$ images per GPU. ResNet-50 trained on the Coco dataset is used for checkpoint initialization. A global step of $1000$ is set for the whole training loop.
\section{Experimental Result}\label{sec7}
In this section, we show the results obtained from selected models on the benchmark dataset used. First, we show their quantities and performances, especially in terms of computational resources, average precision, and average recall for Mask R-CNN \cite{ref_article21} and panoptic quality, precision on small and large targets, and recall quality for sensitivity. Second, we assess the quality and efficiency of selected models. The table \ref{tab2} is divided into two sections: first, the metrics for the Mask R-CNN model, and then the metrics for the Panoptic DeepLab \cite{ref_article24} model. 
\subsection{Quantitative Comparison}\label{51}
 Segmentation methods are preferred in the field of high-resolution recognition due to their mechanisms, including pixel label classification.
\begin{table*}[h]
\begin{center}
\begin{minipage}{\textwidth}
\caption{Table of experimental result}\label{tab2}
\begin{tabular*}{\textwidth}{@{\extracolsep{\fill}}lcccccc@{\extracolsep{\fill}}}
\toprule%
& \multicolumn{3}{@{}c@{}}{Metrics\footnotemark[1]} \\\cmidrule{2-7}%
Model & GPU x Day & NP & AP & APS & APS  & AR\\
\midrule
Mask R-CNN\cite{ref_article21}  & 1.4 & 20.04 M & $35.6$ & - & - & $28.05$ \\\cmidrule{2-7}
Model & GPU x Day & NP & PQ & PS & PL  & RQ\\
Panoptic DeepLab\cite{ref_article24} & 0.5 & 25.6 M & $38.5$ & $37.8$ & $39.1$ & $40,8$\\
\end{tabular*}
\footnotetext{Note: Results are obtained by training on a few samples and do not integrate all data collected and unlabelled}
\footnotetext[1]{Panoptic Quality PQ is a metric corresponding to Average Precision AP for Panoptic Segmentation \cite{ref_article24} }
\end{minipage}
\end{center}
\end{table*}
 Mask R-CNN trained in more than $20.04$ Millions of parameters takes more than $1$ day for execution with Average Precision of $35.6\%$  and a little less on Average Recall. The Panoptic DeepLab model have $+3\%$ more Precision in Panoptic Quality with $25.6$ million parameters trained in a few hours. More or less the same value in small and large objects. We note an increase of $+10\%$ in Average Recall. An analysis of this information is made based on the nature of the models and the environment studied.   

\subsection{Qualitative Comparison}
Instance and panoptic segmentation methods, which include semantic segmentation, have good accuracy on our Litter Beach Dataset with a small amount of data. Inference shows an accurate detection of waste down to a scale of a few centimeters (see Figure 4). The parallel application of the segmentation mask reduces false positives produced by heterogeneity on the coastline. Moreover, Panoptic Deeplab has a higher sensitivity due to the combination of semantic segmentation, which first segments the image by dissociating the background as well as the foreground by recognizing the stuff here as "sand". With this mechanism, object detection is more visible, and this reduces the rate of false negatives. We can see a positive evolution of the semantic loss function as well as a regression with a low rate following center loss on the following curve.

\begin{figure}[H]\label{fig4}
    \begin{minipage}[c]{.46\linewidth}
        \centering
        \includegraphics[width=6cm, height=4.5cm]{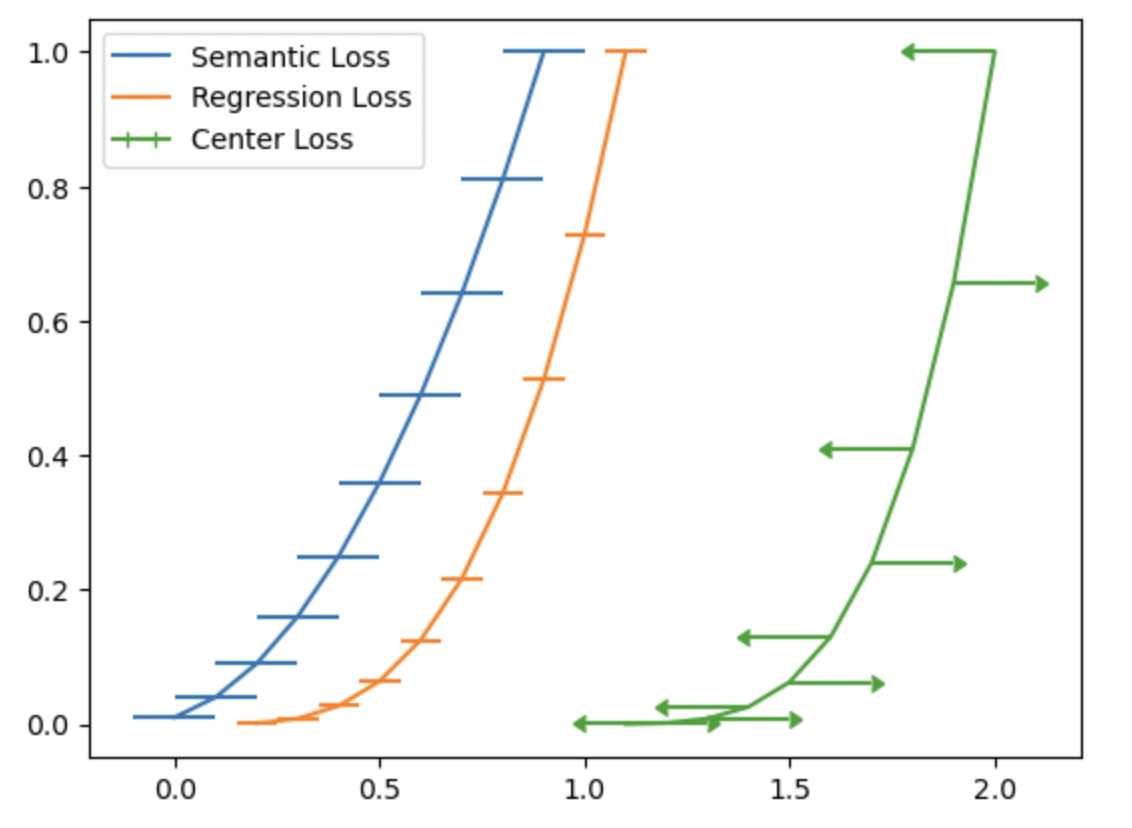}.
    \end{minipage}
    \hfill%
    \begin{minipage}[c]{.46\linewidth}
        \centering
        \includegraphics[width=6cm, height=4.5cm]{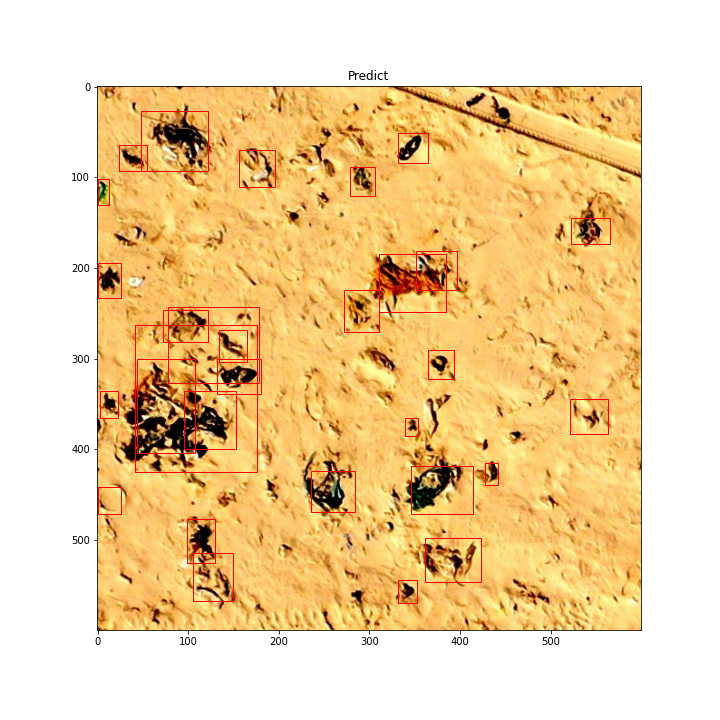}.
    \end{minipage}
    \caption{A) Decreasing of function loss during training step of Panoptic DeepLab, B)Example Prediction of Litter and Algue with MaskRCNN in one images}
\end{figure}

\section{Further Discussions}\label{sec8}
The use of UAV is an important aspect for saving time and having considerable coverage compared to traditional techniques, which are more tedious. The altitude of $10$ meters seems to be more suitable in terms of flight time, coverage, image quality, and waste visibility, as attested by Martins et al. \cite{ref_article4} and Fallati et al. \cite{ref_article7}. The division of the images into several images of adapted sizes to the inputs of our algorithm allows for better visibility of the waste with a zoom effect. To solve the problem, machine learning techniques have been proposed, such as the random forest in \cite{ref_article4} and \cite{ref_article5} with F-scores varying between $40\%$ and $78\%$. It should be noted that in the field of image detection, CNNs are more commonly used than classic methods of machine learning. However, the use of deep learning methods varies depending on the ecosystem studied. Some related work \ref{sec2} use a customized CNN-based model. We opted for the detection of waste on the coast by segmentation methods, including panoptic, for more reliability. The aims of this paper are to first show that the segmentation method is more adaptable in environmental aerial images, such as litter beaches, and then demonstrate the robustness of the model when applying panoptic segmentation.

\section{Conclusion}\label{sec9}
In this work, we showed that it is possible to find marine waste to protect the environment. We have demonstrated that Deep Learning, in particular segmentation methods like Mask R-CNN and Panoptic DeepLab, are adapted on this kind of task and outperforms other classical machine learning methods. This work shows that marine litter detection using Deep learning techniques and UAV drone tools can significantly fight against marine pollution and environmental degradation. So, it is easier for groups that work to protect the ocean to set up an effective system for autonomous monitoring. In other research, it is critical to determine the density of waste elements on the coastline, but it is even more important to have a large scale coverage in accordance with an optimal calculation time. It is also possible to compare the evolution of the states taken through the years.

\section*{Acknowledgment}
This publication was made possible through the DSTN, supported by IRD and AFD. We would like to thank the African Center of Excellence in Mathematical, Informatics, and Tics (CEA-MITIC) and the African Centre of Excellence in Mathematical Science Informatics and their Application (CEA-SMIA) for their support.




%
%
%

\begin{thebibliography}{8}

\bibitem{ref_article1}
Ministères Écologie Énergie Territoires. “Déchets marins.” Accessed January 15, 2023. https://www.ecologie.gouv.fr/dechets-marins.

\bibitem{ref_article2}Velander, Kathy, and Marina Mocogni. 1999. “Beach Litter Sampling Strategies: Is There a ‘Best’Method?” Marine Pollution Bulletin 38 (12): 1134–40.

\bibitem{ref_article3}Wenneker, Barbara, and Lex Oosterbaan. 2010. “Guideline for Monitoring Marine Litter on the Beaches in the OSPAR Maritime Area. Edition 1.0.” Report. OSPAR Commission. https://doi.org/10.25607/OBP-968.

\bibitem{ref_article4}
Martin, Cecilia, Stephen Parkes, Qiannan Zhang, Xiangliang Zhang, Matthew F. McCabe, and Carlos M. Duarte. “Use of Unmanned Aerial Vehicles for Efficient Beach Litter Monitoring.” Marine Pollution Bulletin 131 (2018): 662–73.

\bibitem{ref_article5}
Gonçalves, Gil, Umberto Andriolo, Luís Pinto, and Filipa Bessa. “Mapping Marine Litter Using UAS on a Beach-Dune System: A Multidisciplinary Approach.” Science of the Total Environment 706 (2020): 135742.

\bibitem{ref_article6}
Youme, Ousmane, Theophile Bayet, Jean Marie Dembele, and Christophe Cambier. “Deep Learning and Remote Sensing: Detection of Dumping Waste Using UAV.” Procedia Computer Science, Big Data, IoT, and AI for a Smarter Future, 185 (January 1, 2021): 361–69. https://doi.org/10.1016/j.procs.2021.05.037.

\bibitem{ref_article7}
Fallati, Luca, Annalisa Polidori, Christian Salvatore, Luca Saponari, Alessandra Savini, and P. Galli. “Anthropogenic Marine Debris Assessment with Unmanned Aerial Vehicle Imagery and Deep Learning: A Case Study along the Beaches of the Republic of Maldives.” Science of The Total Environment 693 (2019): 133581.
Geus, Daan de, Panagiotis Meletis, and Gijs Dubbelman. “Fast Panoptic Segmentation Network.” arXiv, October 9, 2019. https://doi.org/10.48550/arXiv.1910.03892.

\bibitem{ref_article8}
Papakonstantinou, Apostolos, Marios Batsaris, Spyros Spondylidis, and Konstantinos Topouzelis. “A Citizen Science Unmanned Aerial System Data Acquisition Protocol and Deep Learning Techniques for the Automatic Detection and Mapping of Marine Litter Concentrations in the Coastal Zone.” Drones 5, no. 1 (2021): 6.

\bibitem{ref_article9}
Simonyan, Karen, and Andrew Zisserman. 2015. “Very Deep Convolutional Networks for Large-Scale Image Recognition.” arXiv. https://doi.org/10.48550/arXiv.1409.1556.

\bibitem{ref_article10}
Wolf, Mattis, Katelijn van den Berg, Shungudzemwoyo P. Garaba, Nina Gnann, Klaus Sattler, Frederic Stahl, and Oliver Zielinski. “Machine Learning for Aquatic Plastic Litter Detection, Classification and Quantification (APLASTIC-Q).” Environmental Research Letters 15, no. 11 (2020): 114042.


\bibitem{ref_article11}
Liao, Yu-Hsien, and Jih-Gau Juang. “Real-Time UAV Trash Monitoring System.” Applied Sciences 12, no. 4 (January 2022): 1838. https://doi.org/10.3390/app12041838

\bibitem{ref_article12}
Redmon, Joseph, Santosh Divvala, Ross Girshick, and Ali Farhadi. 2016. “You Only Look Once: Unified, Real-Time Object Detection.” arXiv. https://doi.org/10.48550/arXiv.1506.02640.


\bibitem{ref_article13}
Chollet, François. “Xception: Deep Learning with Depthwise Separable Convolutions.” arXiv, April 4, 2017. https://doi.org/10.48550/arXiv.1610.02357.



\bibitem{ref_article14}
Chen, Liang-Chieh, George Papandreou, Iasonas Kokkinos, Kevin Murphy, and Alan L. Yuille. “DeepLab: Semantic Image Segmentation with Deep Convolutional Nets, Atrous Convolution, and Fully Connected CRFs.” arXiv, May 11, 2017.

\bibitem{ref_article15}
Ronneberger, Olaf, Philipp Fischer, and Thomas Brox. 2015. “U-Net: Convolutional Networks for Biomedical Image Segmentation.” arXiv. https://doi.org/10.48550/arXiv.1505.04597.

\bibitem{ref_article16}
Long, Jonathan, Evan Shelhamer, and Trevor Darrell. “Fully Convolutional Networks for Semantic Segmentation.” arXiv, March 8, 2015. https://doi.org/10.48550/arXiv.1411.4038.


\bibitem{ref_article17}
Wang, Jingdong, Ke Sun, Tianheng Cheng, Borui Jiang, Chaorui Deng, Yang Zhao, Dong Liu, et al. “Deep High-Resolution Representation Learning for Visual Recognition.” arXiv, March 13, 2020. https://doi.org/10.48550/arXiv.1908.07919.


\bibitem{ref_article18}
Liu, Yinglu, Yan-Ming Zhang, Xu-Yao Zhang, and Cheng-Lin Liu. “Adaptive Spatial Pooling for Image Classification.” Pattern Recognition 55 (July 1, 2016): 58–67. https://doi.org/10.1016/j.patcog.2016.01.030.

\bibitem{ref_article19}
Zhao, Hengshuang, Jianping Shi, Xiaojuan Qi, Xiaogang Wang, and Jiaya Jia. 2017. “Pyramid Scene Parsing Network.” arXiv. https://doi.org/10.48550/arXiv.1612.01105.

\bibitem{ref_article20}
Yang, Tien-Ju, Maxwell D. Collins, Yukun Zhu, Jyh-Jing Hwang, Ting Liu, Xiao Zhang, Vivienne Sze, George Papandreou, and Liang-Chieh Chen. “DeeperLab: Single-Shot Image Parser.” arXiv, March 12, 2019. https://doi.org/10.48550/arXiv.1902.05093.


\bibitem{ref_article21}
He, Kaiming, Georgia Gkioxari, Piotr Dollár, and Ross Girshick. “Mask R-CNN.” arXiv, January 24, 2018. https://doi.org/10.48550/arXiv.1703.06870.


\bibitem{ref_article22}
Tan, Mingxing, and Quoc V. Le. “EfficientNet: Rethinking Model Scaling for Convolutional Neural Networks.” arXiv, September 11, 2020. https://doi.org/10.48550/arXiv.1905.11946.

\bibitem{ref_article23}
Xie, Enze, Peize Sun, Xiaoge Song, Wenhai Wang, Ding Liang, Chunhua Shen, and Ping Luo. “PolarMask: Single Shot Instance Segmentation with Polar Representation.” arXiv, February 25, 2020. https://doi.org/10.48550/arXiv.1909.13226.

\bibitem{ref_article24}
Cheng, Bowen, Maxwell D. Collins, Yukun Zhu, Ting Liu, Thomas S. Huang, Hartwig Adam, and Liang-Chieh Chen. “Panoptic-DeepLab: A Simple, Strong, and Fast Baseline for Bottom-Up Panoptic Segmentation.” arXiv, March 11, 2020. https://doi.org/10.48550/arXiv.1911.10194.
\bibitem{ref_article25}
Mohan, Rohit, and Abhinav Valada. “EfficientPS: Efficient Panoptic Segmentation.” arXiv, February 1, 2021. https://doi.org/10.48550/arXiv.2004.02307.

\bibitem{ref_article26}
Geus, Daan de, Panagiotis Meletis, and Gijs Dubbelman. 2019. “Fast Panoptic Segmentation Network.” arXiv. https://doi.org/10.48550/arXiv.1910.03892.

\bibitem{ref_article27}
Tavares, Davi Castro, Jailson Fulgêncio Moura, Adam Ceesay, and Agostino Merico. “Density and Composition of Surface and Buried Plastic Debris in Beaches of Senegal.” Science of The Total Environment 737 (October 1, 2020): 139633. https://doi.org/10.1016/j.scitotenv.2020.139633.


\bibitem{ref_article28}
Ren, Shaoqing, Kaiming He, Ross Girshick, and Jian Sun. “Faster R-CNN: Towards Real-Time Object Detection with Region Proposal Networks.” arXiv, January 6, 2016. https://doi.org/10.48550/arXiv.1506.01497.

\end{thebibliography}
%

\end{document}